\documentclass[journal]{IEEEtran}
\usepackage{fancyhdr}
\usepackage{times}
\usepackage{epsfig}
\usepackage{graphicx}
\usepackage{amsmath}
\usepackage{amssymb}
\usepackage{algpseudocode}
\usepackage{algorithm} 
\usepackage[breaklinks=true,bookmarks=false]{hyperref}

\begin{document}

	\title{Depth Image Inpainting: Improving Low Rank Matrix Completion with Low Gradient Regularization}
	
	\author{\IEEEauthorblockN{Hongyang Xue,
			Shengming Zhang and
			Deng~Cai,~\IEEEmembership{Member,~IEEE}}
		\IEEEcompsocitemizethanks{
		
			\IEEEcompsocthanksitem H. Xue, S. Zhang, D. Cai are with State Key Lab of CAD\&CG, College of Computer Science,
			Zhejiang University, Hangzhou 310027, China.\protect
			
			E-mail: hyxue@outlook.com, michaelzhang@zju.edu.cn, dengcai@cad.zju.edu.cn
		}
		\thanks{}}
		%
	
		\markboth{IEEE TRANSACTIONS ON. VOL. **, NO. **, APRIL 2016}%
		{Shell \MakeLowercase{\textit{et al.}}:Depth Image Inpainting: Improving Low Rank Matrix Completion with Low Gradient Regularization}
		\maketitle

	\begin{abstract}
		We consider the case of inpainting single depth images. Without corresponding color images, previous or next frames, depth image inpainting is quite challenging. One natural solution is to regard the image as a matrix and adopt the low rank regularization just as inpainting color images. However, the low rank assumption does not make full use of the properties of depth images  .  
		
		A shallow observation may inspire us to penalize  the non-zero gradients by sparse gradient regularization.  However, statistics show that though most pixels have zero gradients, there is still a non-ignorable part of pixels whose gradients are equal to 1.  Based on this specific property of depth images , we propose a low gradient regularization method in which we reduce the penalty for gradient 1 while penalizing the non-zero gradients to allow for gradual depth changes.  The proposed low gradient regularization is integrated with the low rank regularization into the low rank low gradient approach for depth image inpainting. We compare our proposed low gradient regularization with sparse gradient regularization. The experimental results show the effectiveness of our proposed approach.
	\end{abstract}
	

	\section{Introduction}
	
	Image inpainting is an important research topic in the fields of computer vision and image processing\cite{guillemot2014image}, \cite{shi2011non}, \cite{dai2009removing}, \cite{tang2012robust}. A lot of approaches have been proposed to tackle inpainting problems for images of different categories \cite{bertalmio2000image}, \cite{criminisi2004region}, \cite{buades2010image}. However, most research have been focused on natural images and medical images. The research amount on depth image inpainting is relatively small.
	
	The fast development of the RGB-D sensors, such as Microsoft Kinect, ASUS Xtion Pro and Intel Leap Motion, enables a variety of applications based on the depth information by providing depth images of the scenes in real time. Together with the traditional multi-view stereo approaches, depth images are now playing a more and more important role in computer vision research and applications \cite{ren2012rgb}, \cite{rohrbach2012database}, \cite{song2014sliding}, \cite{silberman2012indoor}, yet the inpainting problem of them are not well-studied. 
	
\begin{figure*}
		\label{fig1}
		\begin{minipage}[t]{0.33\linewidth}
			\centering
			\includegraphics[width=2in]{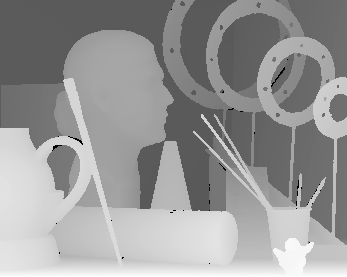}
			\label{fig:side:a}
		\end{minipage}%
		\begin{minipage}[t]{0.33\linewidth}
			\centering
			\includegraphics[width=2in]{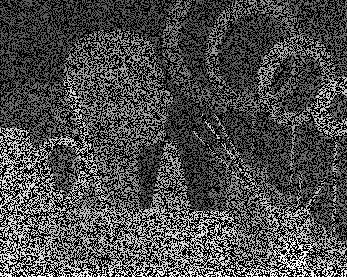}
			\label{fig:side:b}
		\end{minipage}%
		\begin{minipage}[t]{0.33\linewidth}
			\centering
			\includegraphics[width=2in]{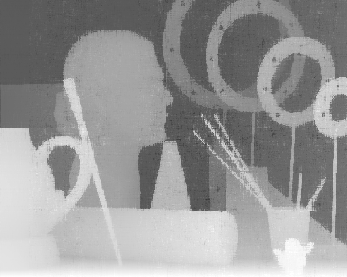}
			\label{fig:side:b}
		\end{minipage}
		\caption{Left: Original disparity map. Middle: Corrupted disparity map with half pixels missing. Right: The inpainted result by low rank(LR) completion. We can see the inpainted map has much noise compared with original map.}
	\end{figure*}

	The main reason may be that most image inpainting techniques can be applied directly to depth images. Noting that there is only a simple mathematical relation between the disparity value and the depth value, we will use disparity instead of depth in our paper. In the remainder of the paper, depth and disparity will have the same meaning which refers to the disparity value.  To obtain coarse inpainting results, we apply the low rank assumption and complete the depth image with the low rank matrix completion approach \cite{candes2009exact}. The inpainting results are not satisfactory enough.  Depth images are textureless compared with natural images. The lack of texture causes difficult for the low rank completion approach. In addition, the low rank completion approach usually results in excessive and spurious details in the inpainted areas (see Figure \ref{fig1}). 
	Moreover, depth images have quite sparse gradients. In other words, gradients vanish at most places. Therefore, together with the textureless property, it is reasonable if one regularizes the inpainting results with the sparse gradient prior. To improve depth inpainting results under the low rank assumption,  the gradients can be regularized in the meanwhile. There have been work in recovering images (e.g. medical images \cite{shi2013low} and natural images \cite{chan2005recent}) under the sparse gradient assumption.


	 However, statistics of depth image gradients show that the sparse gradient assumption is not accurate enough. The gradients can be described more properly as \textit{low} rather than \textit{sparse}. In another word, at many places in the depth image, gradients are not always 0 but rather very small (see Figure \ref{fig2}). This property has not been considered  for image inpainting because it is not universal in general images. For depth images, statistics show the universality of this property. Hence we propose the \textit{low gradient} regularization. We denote the low gradient regularization as  $L_0^\psi$ gradient regularization. The notation comes from the corresponding non-convex TV$^\psi$ norm \cite{hintermuller2014superlinearly}. Relation between them will be explained in the remainder part of this paper.
	
	Unlike the $L_0$ norm which penalizes the non-zero elements equally (the norm is always 1 whether the element is 1 or 100), our proposed $L_0^\psi$ measure reduces the penalty for small elements. In depth images, our $L_0^\psi$ reduces the penalty for gradient 1 (all gradients are truncated into integer values) and thus allows for gradual depth changes.
	
	Our main contributions are two-folds: First we propose the low gradient regularization $L_0^\psi$ which well describes the statistical property of the depth image gradients. Second we develop a  solution to the $L_0^\psi$ gradient minimization problem based on \cite{nguyen2015fast}.  We integrate our low gradient regularization with the low rank assumption into the LRL0$^\psi$ regularization  for depth image inpainting. In the experiments we compare our LRL0$^\psi$ algorithm with two approaches, the low rank total variation (LRTV) \cite{shi2013low} and the low rank L0 gradient (LRL0) approaches, which only enforce the sparse gradient constraint.
	
	The remainder of the paper is organized as follows. Section \ref{sec2} describes related work in image inpainting and the sparse gradient.  Section \ref{sec3} introduces the LRTV \cite{shi2013low} and the LRL0 approaches which only consider the sparse gradient regularization. In section \ref{sec3} we point out the defects of the sparse gradient regularization and then our main contribution, the low gradient regularization, is described in section \ref{sec4}. In section \ref{sec5} we perform experiments on our dataset and display the experimental results. Finally We conclude our work in \ref{sec6}.
	
	\section{Background} 
	\label{sec2}
	\subsection{Low Rank Matrix Completion}
	Completing a matrix with missing observations is an intriguing task in the machine learning and mathematics society \cite{cai2010singular}, \cite{candes2010power}, \cite{candes2010matrix}, \cite{keshavan2009matrix}, \cite{zheng2012practical}. Most work are based on the low rank assumption of the underlying data. The low rank matrix completion is one of effective approaches for image inpainting. Candes \textit{et al.} \cite{candes2009exact} first introduce the matrix completion problem by approximating the rank with nuclear norm. Zhang \textit{et al.} \cite{zhang2012matrix} propose the truncated nuclear norm regularization and achieve excellent results in image inpainting. Gu \textit{et al.} \cite{gu2014weighted} further present the weighted nuclear norm regularization and perform the image denoising task with outstanding results.
	
	\subsection{Depth Inpainting}
	
	Depth image inpainting has been considered mostly under the situation of RGB-D inpainting or stereoscopic inpainting problems. Moreover, most depth inpainting approaches inpaint missing regions of specific kinds (e.g. occlusions, missing caused by sensor defects or holes caused by object-removal). Doria \textit{et al.} \cite{doria2012filling} introduce a technique to fill holes in the LiDAR data sets. Wang \textit{et al.} \cite{wang2008stereoscopic} present an algorithm for simultaneous color and depth inpainting. They take the stereo image pairs and the estimated disparity map as input and fill the holes cause by object removal. Lu \textit{et al.} \cite{Lu_2014_CVPR} cluster the RGB-D image patches into groups and employ the low rank matrix completion approach to enhance the depth images obtained from  RGB-D sensors with the aid of corresponding color images. Zou \textit{et al.} \cite{zou2014automatic} consider inpainting RGB-D images by removing fence-like structures. Buyssens \textit{et al.} \cite{buyssens2015superpixel} inpaint holes in the depth maps based on superpixel for virtual view synthesizing in RGB-D scenes. Herrera \textit{et al.} \cite{herrera2013depth} inpaint incomplete depth maps produced by 3D reconstruction methods under a second-order smoothness prior. As far as we know, almost all depth inpainting approaches refer to color images. They are either color-guided depth inpainting or simultaneous RGB-D inpainting. However, we consider inpainting only a single depth image.
	
	\subsection{TV and $L_0$ Gradient Regularization}
	TV norm and its variations have been employed in image processing for quite a long time \cite{chan2005recent}, \cite{goldluecke2010approach}. Perrone \textit{et al.} \cite{perrone2014total} propose a blind deconvolution algorithm based on the total variation minimization. Guo \textit{et al.} \cite{guo2015generalized} extend the total variation norm to tensors for visual data recovery. Li \textit{et al.} \cite{li2014non} use the total variation in their robust noisy image completion algorithm. 
	
	Total variation refers to the integration of the norm of gradients on the whole image. When a function is applied to the norm of the gradients at each pixel before integral, the integration is called the total generalized variation. Hinterm{\"u}ller \textit{et al.} \cite{hintermüller2013nonconvex} propose a nonconvex TV$^q$ model for image restoration. They further propose a superlinearly solver for the general concave generalized TV model in \cite{hintermuller2014superlinearly}. Ranftl \textit{et al.} \cite{ranftl2014non} utilize the total generalized variation for optical flow estimation.
	
	As mentioned above, the TV norm is a relaxation of the $L_0$ gradient. However, the TV norm also penalizes large gradient magnitudes. It may influence real image edges and boundaries \cite{xu2011image}. Thus many algorithms directly solving $L_0$ gradient minimization have been proposed. Xu \textit{et al.} \cite{xu2011image} adopt a special alternating optimization strategy. They employ the method for image deblurring \cite{xu2013unnatural}. Their work is later applied for mesh denoising by He \textit{et al.} \cite{he2013mesh}. Nguyen \textit{et al.} \cite{nguyen2015fast} propose a fast $L_0$ gradient minimization approach based on a method named region fusion. 
	
	There are approaches that combine low rank regularization with the total variation minimization. Shi \textit{et al.} \cite{shi2013low} 
	combine both the low rank and the total variation regularization into an LRTV algorithm for medical image super-resolution. Ji \textit{et al.} \cite{ji2016tensor} propose a tensor completion approach based on the total variation and the low-rank matrix completion. He \textit{et al.} \cite{he2016total} employ the total variation regularization and the low-rank matrix factorization for hyperspectral image restoration.

	\section{Low Rank Sparse Gradient Approach}
	\label{sec3}
	In this section, we will review the LRTV  \cite{shi2013low} algorithm and introduce how it can be employed to inpaint depth images.  Then the $L_0$ gradient is used to replace the TV regularization. We briefly review the algorithm developed by Nguyen \textit{et al.} \cite{nguyen2015fast} for the $L_0$ gradient minimization. These are necessary for the explanation of our main contribution, the LRL0$^\psi$ algorithm,  which will be described in the next section.
	
	Given a corrupted disparity map or depth image $\mathbf{D}$ and its inpainting mask $\mathbf{\Omega}$ (missing areas), we hope to recover the original image $\mathbf{U}$. The recovered image $\mathbf{U}$ should match the observations $\mathbf{U}|_\Omega = \mathbf{D}|_\Omega$. In conventional matrix completion image inpainting scheme, the low rank prior is added as a regularization \cite{candes2009exact}, \cite{gu2014weighted}, \cite{zhang2012matrix} and the unknown image is recovered by solving 
	
	\begin{equation}\label{eq1}
		\arg\min_\mathbf{U}||\mathbf{U}-\mathbf{D}||^2_\Omega + \lambda \cdot \mathit{rank}(\mathbf{U})
	\end{equation}
	
	$\lambda$ is a weight representing the importance of low rank. The above rank minimization problem is intractable due to the non-convexity and discontinuous nature of the rank function. Theoretical studies show that the nuclear norm is the tightest convex lower bound of the rank function of matrices \cite{recht2010guaranteed}. Therefore, rank is usually approximated by the nuclear norm \cite{candes2009exact}. We employ the nuclear norm  \cite{candes2009exact} as the low rank prior. As discussed in the introduction, we also add the sparse gradient regularization for depth recovery. Altogether we have the following formula
	
	\begin{equation}\label{eq2}
		\arg\min_\mathbf{U}||\mathbf{U}-\mathbf{D}||^2_\Omega + \lambda_r \cdot ||\mathbf{U}||_{*} + \lambda_s \cdot ||\nabla \mathbf{U}||_0
	\end{equation} 
	
	Notice that the third term in equation \ref{eq2} is the $L_0$ norm of gradient. Minimization corresponding to the $L_0$ norm is usually relaxed to $L_1$ norm and thus the $L_0$ gradient becomes total variation \cite{shi2013low}, \cite{he2016total}, \cite{ji2016tensor}. The $L_0$ norm is non-convex. The advantage of relaxing the $L_0$ gradient to total variation is that the problem becomes convex. We first employ the existed LRTV scheme as described in \cite{shi2013low} for depth inpainting.
	\subsection{Total Variation}
	
	The $L_0$ gradient norm is approximated by total variation and equation \ref{eq2} now becomes
	
	\begin{equation}\label{eq3}
		\arg\min_\mathbf{U}||\mathbf{U}-\mathbf{D}||^2_\Omega + \lambda_r \cdot ||\mathbf{U}||_{*} + \lambda_{tv} \cdot TV(\mathbf{U})
	\end{equation}
	
	This problem has almost the same form as the super-resolution problem in \cite{shi2013low}. Following the solution in \cite{shi2013low}, we employ the ADMM alrgorithm \cite{stephen2011distributed} to solve the new equation
	
	\begin{equation}\label{eq4}
		\begin{split}
			min_{\mathbf{U},\mathbf{M},\mathbf{Y}}||&\mathbf{U}-\mathbf{D}||^2_\Omega + \lambda_r||\mathbf{M}||_{*} + \lambda_{tv}TV(\mathbf{U}) + \\ \frac{\rho}{2}&(||\mathbf{U}-\mathbf{M}+\mathbf{Y}||^2-||\mathbf{Y}||^2)
		\end{split}
	\end{equation}
	
	The minimization problem in equation \ref{eq4} is broken into three sub-problems and the variables are iteratively updated.
	
	The first subproblem needs to update $\mathbf{U}^{k+1}$ by minimizing part of equation \ref{eq4} related with total variation.
	\begin{equation}\label{eq5}
		\arg\min_{\mathbf{U}}||\mathbf{U}-\mathbf{D}||^2_\Omega + \lambda_{tv}TV(\mathbf{U}) + \frac{\rho}{2}||\mathbf{U}-\mathbf{M}^k+\mathbf{Y}^k||^2
	\end{equation}
	
	This subproblem is solved by Bregman iteration \cite{marquina2008image}.
	In the second subproblem $\mathbf{M}^{k+1}$ is updated by
	\begin{equation}\label{eq6}
		\arg\min_{\mathbf{M}}||\mathbf{M}||_{*}+\frac{\rho}{2}||\mathbf{U}^{k+1}-\mathbf{M}+\mathbf{Y}^k||^2
	\end{equation}

	After the update of $\mathbf{U}^{k+1}$ and $\mathbf{M}^{k+1}$, $\mathbf{Y}^{k+1}$ is updated by $\mathbf{Y}^{k+1} = \mathbf{Y}^k + (\mathbf{U}^{k+1} - \mathbf{M}^{k+1})$.
	
	In our experiments, we initialize $\mathbf{U}$ as the coarse inpainting results obtained by the low rank matrix completion and $\mathbf{M}$ and $\mathbf{Y}$ are set to $\mathbf{0}$.
	
	\begin{figure*}
		\centering
		\begin{minipage}[t]{0.33\linewidth}
			\centering
			\includegraphics[width=2in]{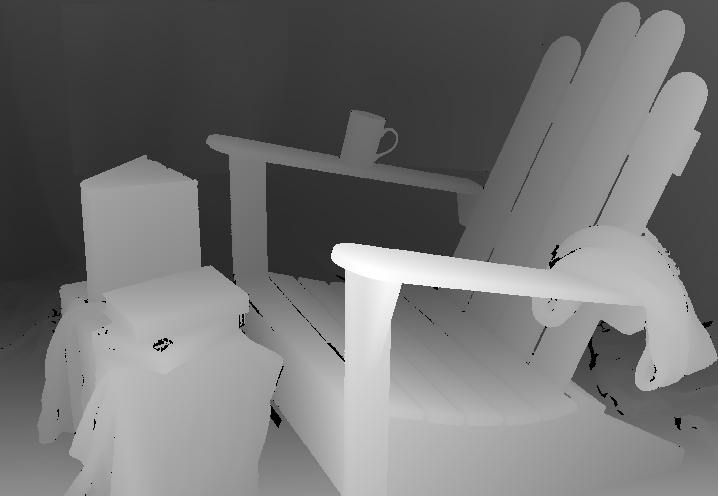}
			\label{fig1:side:a}
		\end{minipage}%
		\begin{minipage}[t]{0.33\linewidth}
			\centering
			\includegraphics[width=2in]{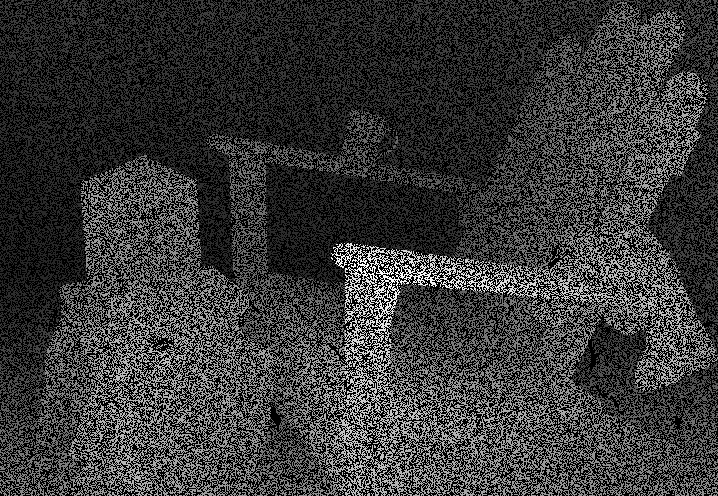}
			\label{fig1:side:d}
		\end{minipage}
		
		\begin{minipage}[t]{0.33\linewidth}
			\centering
			\includegraphics[width=2in]{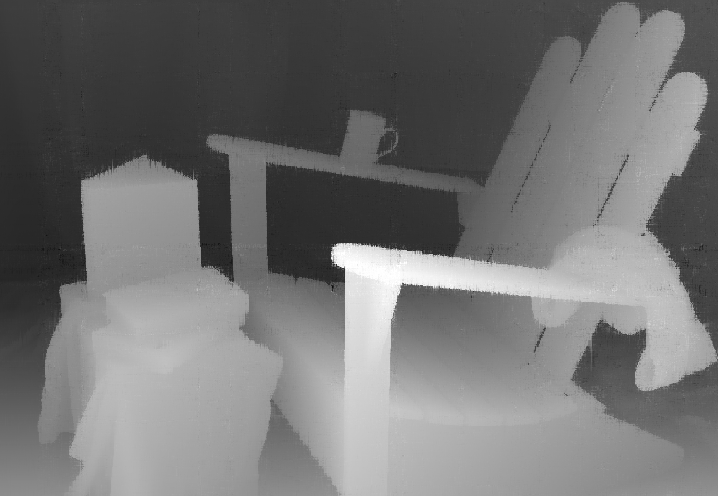}
			\label{fig1:side:e}
		\end{minipage}%
		\begin{minipage}[t]{0.33\linewidth}
			\centering
			\includegraphics[width=2in]{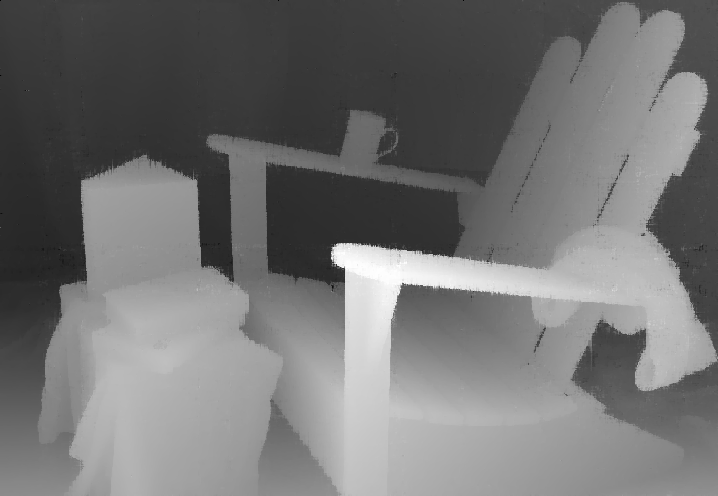}
			\label{fig1:side:c}
		\end{minipage}%
		\caption{Top left: Groundtruth (GT). Top right: Corrupted with 50\% missing entries. Lower left: Low rank inpainting \cite{candes2009exact}. Its PSNR = 27.7596. Lower right: LRTV result. Its PSNR = 28.1319. The LRTV approach obtains better results than only low rank. We also compute the TV norm for GT, low rank, and LRTV. They are 642186, 833767 and 622276 respectively. We can see the LRTV result has lower TV than GT. The $L_0$ norm of them are 63765, 117922 and 81644. The LRTV result has much higher $L_0$ norm than GT. In term of TV regularization, LRTV has achieved its optimal solution but in $L_0$ norm it is still far from optimal. The minimization of the TV norm does not necessarily mean the minimization of $L_0$ gradients.}
		\label{fig3}
	\end{figure*}
	
	The TV regularization improves the inpainting results compared with only low rank \cite{candes2009exact} (Figure \ref{fig3}). However, TV regularization has some drawbacks, it smoothes real depth edges \cite{xu2011image}.  We also observe that in depth inpainting results, the TV norm always becomes close and even lower than that of the groundtruth (see Figure \ref{fig3}). And even with a \textit{lower-than-groundtruth} TV, the depth image remains noisy visually. We observe that even we optimize the results to have \textit{lower-than-groundtruth} TV norm, the $L_0$ norm of gradients is still far above the groundtruth. Because the optimal solution under the TV norm regularization is not exactly optimal for the $L_0$ gradient , we decide to directly employ the $L_0$ gradient regularization.
	
	In theory, the $L_0$ norm is the most suitable measure for sparsity. There are research on the gap between TV and $L_0$ (e.g. \cite{pang2015improved}, \cite{candes2008enhancing}). There also have been work on directly solving $L_0$ gradient minimization problems \cite{xu2011image}, \cite{nguyen2015fast} by approximation strategies.
	
	\subsection{$L_0$ Gradient}\label{sec1}
	
	As mentioned above, the TV norm is widely used as the approximation of the $L_0$ norm of gradients for it enables fast and tractable solutions. However, the drawbacks of this approximation are also studied \cite{xu2011image}. Therefore, a lot of approaches which directly and approximately minimize the $L_0$ gradient have been proposed \cite{cheng2014feature}, \cite{xu2011image}, \cite{storath2014jump}, \cite{nguyen2015fast}.
	Among the approaches, Nguyen \textit{et al.} \cite{nguyen2015fast} propose a region fusion method for $L_0$ gradient minimization. Their approach achieves rather good results while running the most efficiently compared with other $L_0$ gradient minimization algorithms \cite{cheng2014feature}, \cite{xu2011image}, \cite{storath2014jump}.
	
	We replace the TV norm in subproblem 2 (equation \ref{eq5}) with $L_0$ gradient
	
	\begin{equation}\label{eq7}
		\arg\min_{\mathbf{U}}||\mathbf{U}-\mathbf{D}||^2_\Omega + \lambda_{l0}||\mathbf{\nabla U}||_0 + \frac{\rho}{2}||\mathbf{U}-\mathbf{M}^k+\mathbf{Y}^k||^2
	\end{equation}
	
	Following \cite{nguyen2015fast}, we rewrite equation \ref{eq7} as 
	
	\begin{equation}\label{eq8}
		\begin{split}
			\arg\min_{\mathbf{U}}\sum_{i=1}^L||U_i-D_i||^2_\Omega +&  \frac{\rho}{2}||U_i-M^k_i+Y^k_i||^2 \\ +& 
			\frac{\lambda_{l0}}{2}\sum_{j\in N_i}||U_i-U_j||_0 
		\end{split}
	\end{equation}
	where $L$ is the length of the signal (the number of pixels in the image) and $N_i$ the neighboring set (four connected pixels) of the $i^{th}$ pixel.
	
	The bold symbol like $\mathbf{U}$ indicates matrix and the normal symbol $U_i$, $M_i$, $D_i$ and $Y_i$ denote the values of $\mathbf{U}$, $\mathbf{M}$, $\mathbf{D}$ and $\mathbf{Y}$ at the $i$th pixel or the mean value in the $i$th region.
	
	In \cite{nguyen2015fast} they propose an algorithm which loops though all neighboring regions (groups) $G_i$ and $G_j$. At first, all pixels are themselves groups. For region $G_i$, $D_i$, $M_i$ and $Y_i$ denote the mean value of $\mathbf{G}$, $\mathbf{D}$ and $\mathbf{Y}$ in the $i$th region $G_i$. We rewrite the pairwise region cost for our objective function as
	\begin{equation}\label{eq9}
		\begin{split}
			f = \min_{U_i,U_j}w_i||U_i-D_i||_\Omega^2 +& w_j||U_j - D_j||_\Omega^2  \\+& \beta c_{i,j}||U_i-U_j||_0\\
			+& \frac{\rho w_i}{2}||U_i - M^k_i + Y^k_i||^2 \\ +& \frac{\rho w_j}{2}||U_j - M^k_j + Y^k_j||^2
		\end{split}
	\end{equation}
	where $\beta$ is the auxiliary parameter ($0\le \beta \le \lambda$) \cite{nguyen2015fast}.
	
	The \textit{fusion criterion} \cite{nguyen2015fast} derived from equation \ref{eq9} which decides whether region $G_i$ and $G_j$ should be fused now becomes
	
	\begin{equation}\label{eq10}
		\{U_i,U_j\} = 
		\begin{cases}
			\{A,A\} & \mathtt{if} \; f_A\le f_B\\
			\{B_i,B_j\} & \mathtt{otherwise}
		\end{cases}
	\end{equation}
	where $A = (\tilde{w}_iD_i+\tilde{w}_jD_j + \rho w_i M^k_i + \rho w_j M^k_j-\rho w_i Y^k_i - \rho w_j Y^k_j) / (\tilde{w}_i + \tilde{w}_j + \rho w_i + \rho w_j)$, $B_i = (2\tilde{w}_iD_i+\rho w_i(M^k_i-Y^k_i))/(2\tilde{w}_i+\rho w_i)$, $B_j = (2\tilde{w}_jD_j+\rho  w_j(M^k_j-Y^k_j))/(2\tilde{w}_j+\rho w_j)$, $f_A$ is the value of equation \ref{eq9} when $U_i = U_j = A$. $f_B$ is the value of equation \ref{eq9} when $U_i = B_i$ and $U_j = B_j$. $w_i$ is the total number of pixels in region $i$ while $\tilde{w}_i$ only counts pixels in region $i$ that are meanwhile \textbf{not} in the missing region $\Omega$.
	
	Then we modifty the region fusion minimization algorithm in \cite{nguyen2015fast} to solve the equation \ref{eq8}. In their original region fusion minimization, the two regions will remain untouched if the criterion equation \ref{eq10} judges they should not be fused. In our settings, we will update them by $U_i = B_i$ and $U_j = B_j$.
	
\begin{figure*}[t]
		\centering
		\begin{minipage}[t]{0.25\linewidth}
			\centering
			\includegraphics[width=1.5in]{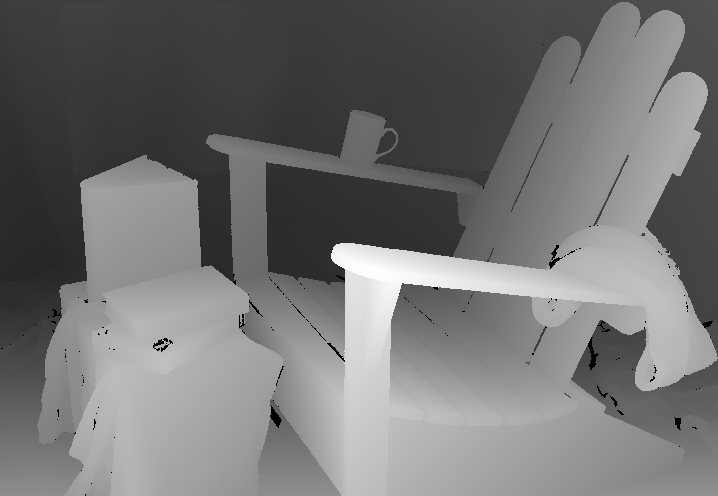}
			\label{fig4:side:b1}
		\end{minipage}%
		\begin{minipage}[t]{0.25\linewidth}
			\centering
			\includegraphics[width=1.5in]{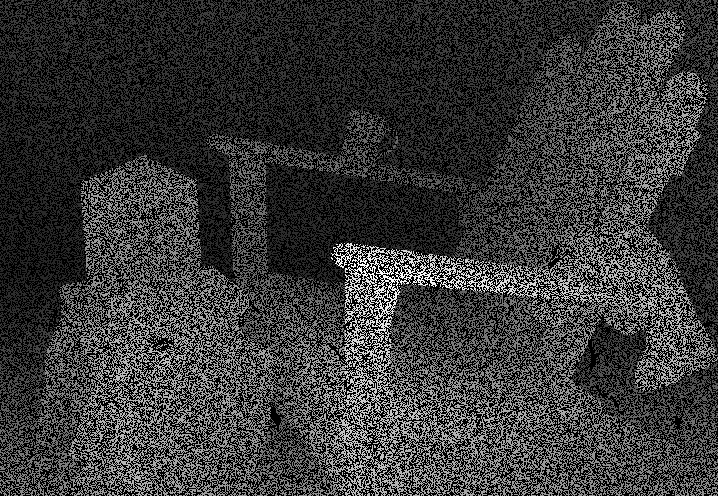}
			\label{fig4:side:c1}
		\end{minipage}%
		\begin{minipage}[t]{0.25\linewidth}
			\centering
			\includegraphics[width=1.5in]{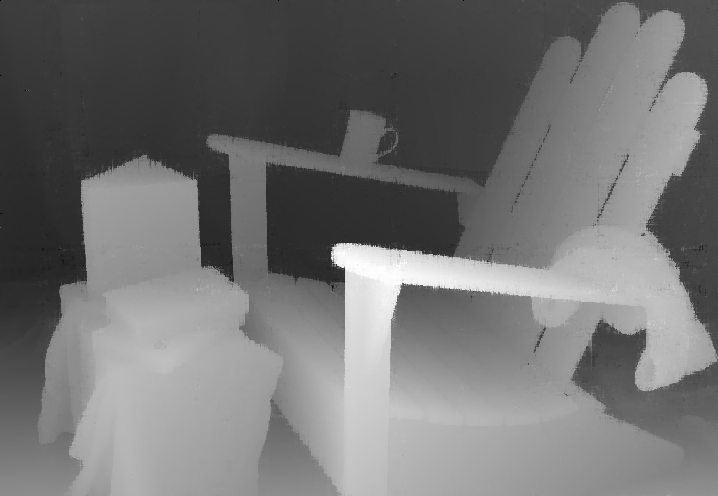}
			\label{fig4:side:d1}
		\end{minipage}%
		\begin{minipage}[t]{0.25\linewidth}
			\centering
			\includegraphics[width=1.5in]{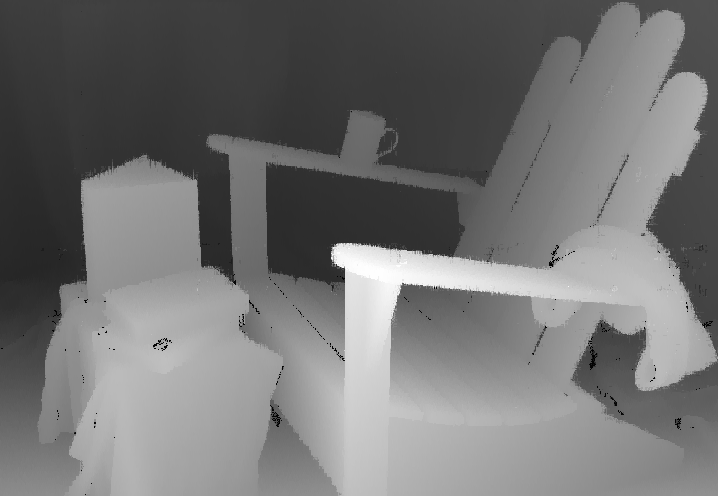}
			\label{fig4:side:e1}
		\end{minipage}
		\hfill
		\begin{minipage}[t]{0.25\linewidth}
			\centering
			\includegraphics[width=1.5in]{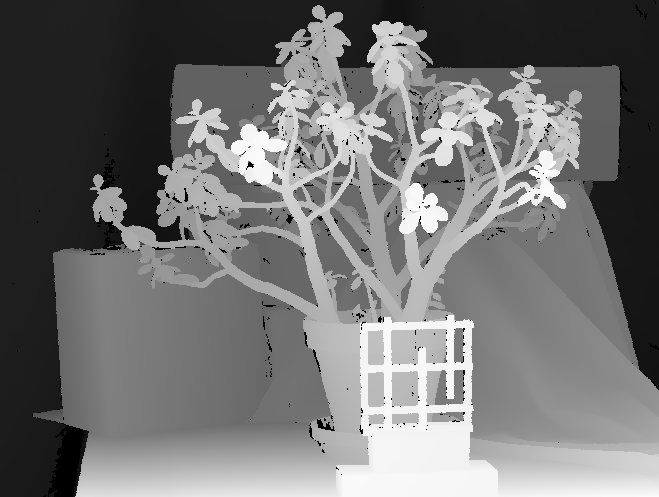}
			\label{fig4:side:a2}
		\end{minipage}%
		\begin{minipage}[t]{0.25\linewidth}
			\centering
			\includegraphics[width=1.5in]{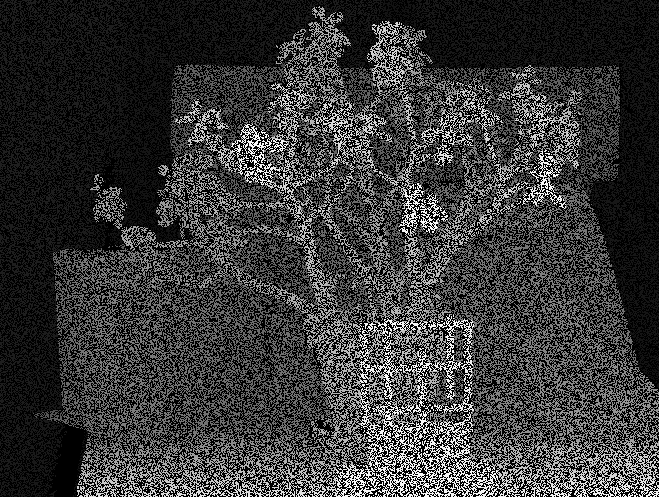}
			\label{fig4:side:b2}
		\end{minipage}%
		\begin{minipage}[t]{0.25\linewidth}
			\centering
			\includegraphics[width=1.5in]{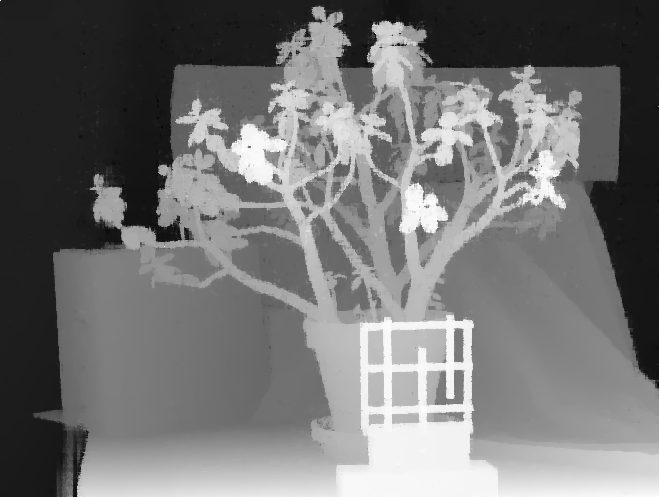}
			\label{fig4:side:c2}
		\end{minipage}%
		\begin{minipage}[t]{0.25\linewidth}
			\centering
			\includegraphics[width=1.5in]{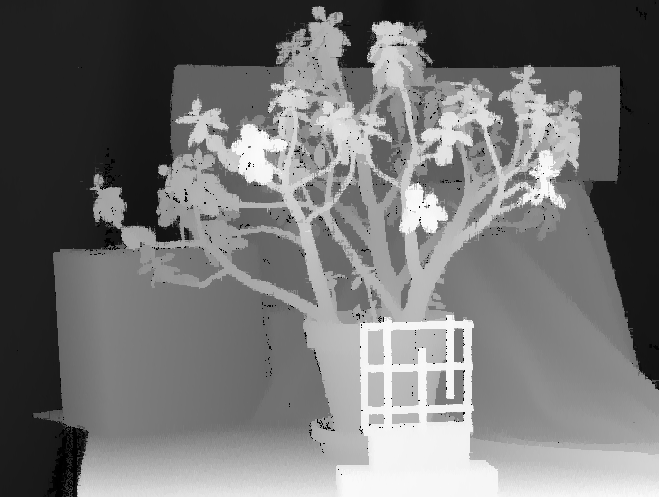}
			\label{fig4:side:e2}
		\end{minipage}

		\caption{The first column are original depth images. The second column corresponds to the corrupted depth images. The third column displays the LRTV results. The last column shows our LRL0 results. For first row (Adirondack), the PSNR of LRTV and LRL0 are 28.1319 and 28.6198 respectively. For second row (Jadeplant), they are 22.8231 and 22.7730. We can see LRL0 achieves better inpainting results than LRTV on Adirondack but fails on Jadeplant.}
		\label{fig4}
	\end{figure*}

	The $L_0$ gradient regularization leads to better results in depth inpainting results in most cases. However, it does not always perform better than LRTV (see Figure \ref{fig4}). One of the reasons may be that the region fusion solution for $L_0$ gradient minimization is only an approximation. But beyond that, we have not fully utilized the features of the gradient maps. We compute the gradients of depth images and find out that the low $L_0$ norm is not accurate enough to characterize the property of depth image gradients (As shown in Figure \ref{fig2}). Besides 0, a non-ignorable part of pixels have gradient 1. In $L_0$ norm, all gradients larger than 0 are penalized equally. Based on the statistics of depth images, we hope to reduce the penalty for gradient 1 so that gradual depth change is allowed.
		
	\section{Low Rank Low Gradient Approach}
	\label{sec4}
	In this section, we will describe our main contribution, the LRL0$^\psi$ algorithm. First we will point out our definition of the low gradient. Then we will describe our low rank low gradient regularization algorithm.
	
		\subsection{Integral Gradient}
		The magnitude of gradient is usually a real number. However, we will consider the gradients as integral values.
		We deal with depth images and disparity images with integral depth values. Thus the gradients on each direction take integral values. When doing statistics on the depth gradients, we truncate the magnitude of the  gradients into integers. Denote the gradient as $(\nabla_x, \nabla_y)$, the truncation of the gradient magnitude is $\biggl \lfloor \sqrt{\nabla_x^2+\nabla_y^2} \biggr \rfloor$. Noting that the truncated value takes 0 only and if only the gradients on both directions are 0. The truncated gradient magnitude takes 1 if and only if the gradients are $(\pm 1, \pm 1)$, $(0, \pm 1)$ or $(\pm 1, 0)$. In other words, for zero gradients, the real values and integral values are exactly the same. For gradient of value 1, it relates to 8 patterns $(\pm 1, \pm 1)$, $(0, \pm 1)$ or $(\pm 1, 0)$. Our low gradient is defined on the integral gradients.
		 	\begin{figure}[t]
		 		\begin{center}
		 			\includegraphics[width=0.8\linewidth]{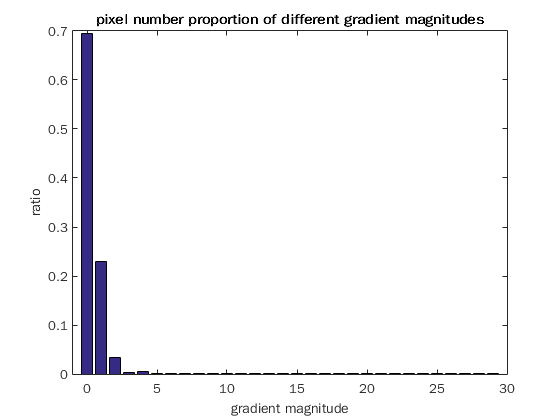}
		 		\end{center}
		 		\caption{Gradient magnitude histogram of groundtruth disparity map of Middlebury Adirondack dataset. We can see most pixels have gradient magnitude 0 and a non-ignorable part have magnitude 1. This observation inspires us to employ \textit{low gradient} regularization.}
		 		\label{fig2}
		 	\end{figure}
		\subsection{Low Gradient}
	As discussed in previous sections, the gradients of depth images cannot be simply depicted as sparse. The penalty of the small gradient value 1 should be reduced to allow for gradual depth changes (see Figure \ref{fig2}). There is a category of generalized TV$^\psi$ norm \cite{hintermuller2014superlinearly} where the penalty of gradients are not increased linearly. 
	Similar to the generalization of TV, we propose the $L_0^\psi$ \textit{norm} so that the penalty for gradient 1 is reduced. Actually it is not a norm but rather a \textit{measure} for the property of data because it does not satisfy the absolute homogeneity. Therefore we will call  $L_0^\psi$ a measure. The $L_0^\psi$ measure is as follows:
	
	\begin{equation}\label{eq11}
		||\mathbf{X}||_{L_0^\psi} = \alpha \#\{X=1\} + \#\{X > 1\}
	\end{equation}
	where $0 < \alpha < 1$ and $\#\{\cdot\}$ denotes the number of elements in the set.
	
	We set $\alpha = 0.75$ in all our experiments based on the statistics of groundtruth gradients.
	
	Thus our low gradient leads to the following optimization problem:
	
	\begin{equation}
			\arg\min_{\mathbf{U}}||\mathbf{U}-\mathbf{D}||^2_\Omega + \lambda_{l0^\psi}||\mathbf{\nabla U}||_{l0^\psi} + \frac{\rho}{2}||\mathbf{U}-\mathbf{M}^k+\mathbf{Y}^k||^2
	\end{equation}
	where $\lambda_{l0^\psi}$ is the weight of importance of the $L_0^\psi$ gradient term.
	
	We extend the region fusion minimization in section \ref{sec1}  to solve $L_0^\psi$ gradient minimization problem.
	
	In this case, the pairwise cost is as follows:
	\begin{equation}\label{eq12}
		\begin{split}
			f = \min_{U_i,U_j}w_i||U_i-&D_i||_\Omega^2 + w_j||U_j - D_j||_\Omega^2  \\+& \beta c_{i,j}||U_i-U_j||_{L_0^\psi}\\
			+& \frac{\rho w_i}{2}||U_i - M^k_i + Y^k_i||^2 \\ +& \frac{\rho w_j}{2}||U_j - M^k_j + Y^k_j||^2
		\end{split}
	\end{equation}
	The fusion criterion now contains three conditions. For groups $G_i$ and $G_j$,
	
	\begin{itemize}
		\item $U_i = U_j$. The optimal solution is $U_i = U_j = A$, where 
		{
			\begin{equation}\label{eq13}
				\begin{split}
					A &= \frac{A_1}{A_2}\\
					A_1 &= 2(\tilde{w}_iD_i+\tilde{w}_jD_j) +\rho w_i(M_i^k-Y_i^k) \\&+\rho w_j (M_j^k-Y^k_j)
					\\
					A_2 &= {2\tilde{w}_i+2\tilde{w}_j+\rho w_i+\rho w_j}
				\end{split}
			\end{equation}
		}
		
		The function value of equation \ref{eq12} under this condition is $f_A$.
		
		\item $|U_i-U_j| = 1$, that is, $U_j = U_i \pm 1$. The optimal solution is $U_i = B$, $U_j = B \pm 1$, where
		
		\begin{equation}\label{eq14}
			\begin{split}
				B &= \frac{B_1}{A_2} \\
				B_1 &= 2\tilde{w}_iD_i + 2\tilde{w}_j(D_j\mp 1)
				\\ &+ \rho w_i(M^k_i-Y^k_i) \\ &+ \rho w_j(M^k_j-Y^k_j\mp 1)
			\end{split}
		\end{equation}
		
		Denote the function value of equation \ref{eq12} under this condition $f_B$.
		\item $|U_i-U_j| > 1$. In this case $U_i = C_i$, $U_j = C_j$. $C_i = (2\tilde{w}_iD_i+\rho w_i(M^k_i-Y^k_i))/(2\tilde{w}_i+\rho w_i)$, $C_j = (2\tilde{w}_jD_j+\rho  w_j(M^k_j-Y^k_j))/(2\tilde{w}_j+\rho w_j)$. We denote the function value as $f_C$.
		
		The \textit{fusion criterion} now becomes:
		
		\begin{equation}\label{eq15}
			\{U_i,U_j\} = 
			\begin{cases}
				\{A,A\} & \mathtt{if} \; f_A\le f_B\; \mathtt{and} \;f_A \le f_C\\
				\{B,B\pm 1\} & \mathtt{if} \; f_B < f_A \; \mathtt{and}\; f_B \le f_C \\
				\{C_i, C_j\} & \mathtt{otherwise}
			\end{cases}
		\end{equation}
	\end{itemize}
	
	Based on this \textit{fusion criterion}, we modify the region fusion iterations in \cite{nguyen2015fast} to solve the $L_0^\psi$ gradient minimization problem (see Algorithm~\ref{alg:the_alg}). Our region fusion low gradient minimization algorithm is summarized in Algorithm \ref{alg:the_alg}. Notice that compared with the region fusion minimization algorithm for $L_0$ in \cite{nguyen2015fast}, our algorithm has several differences. Similar to the modification in $L_0$ gradient minimization, when two regions are not to be fused by the \textit{fusion criterion}, we update them to the optimal solutions $\{B,B\pm 1\}$ or $\{C_i,C_j\}$ following equation \ref{eq15}.

		\begin{algorithm}[htbp]		
		\label{alg}
				\caption{Region Fusion Minimization for $L_0^\psi$}
			\begin{algorithmic}[1] 
			\label{alg:the_alg} 
				\Require
			
				image $U$ with pixel number $N$, the level of sparseness $\lambda$, missing region $\Omega$, original missing image $D$, $Y$ and $M$
				\State Initialize the regions as pixels themselves $G_i\gets \{i\}$, $V_i\gets C_i$, $w_i\gets 1$
				\State $\tilde{w}_i\gets 0$ if $i\in \Omega$, otherwise $\tilde{w}_i \gets 1$
				\State Set $N_i$ as the four-connected neighborhood of $i$
				\State Set $c_{i,j}=1$ if $j\in N_i$, otherwise $c_{i,j} = 0$
				\State $\beta \gets 0$, $iter \gets 0$, $P\gets N$
				\Repeat
				\State $i\gets 1$
				\While{$i\le P$}
				\ForAll{$j\in N_i$}
				\State Compute $f_A,f_B,f_C$ following Section \ref{sec4}
				\If{$f_A\le f_B$ and $f_A\le f_C$}
				\State $G_i\gets G_i\cup G_j$
				\State $V_i\gets (w_iV_i+w_jV_j)/(w_i+w_j)$
				\State $w_i\gets w_i+w_j$
				\State $\tilde{w}_i\gets \tilde{w}_i+\tilde{w}_j$
				\State Remove $j$ in $N_i$ and delete $c_{i,j}$
				\ForAll{$k\in N_j \backslash \{i\}$}
				\If{$k\in N_i$}
				\State $c_{i,k}\gets c_{i,k} + c_{j,k}$
				\State $c_{k,i}\gets c_{i,k} + c_{j,k}$
				\Else
				\State $N_i\gets N_i\cup\{k\}$
				\State $N_k\gets N_k\cup  \{i\}$
				\State $c_{i,k}\gets c_{j,k}$
				\State $c_{k,i}\gets c_{j,k}$
				\EndIf
				\State Remove $j$ in $N_k$ and delete $c_{k,j}$
				\EndFor
				\State Delete $G_j$,$N_j$,$w_j$
				
				\ElsIf{$f_B < f_A$ and $f_B \le f_C$}
				\If{$U_i>U_j$}
				\State $V_i\gets B$, $V_j \gets B-1$
				\Else
				\State $V_i\gets B$, $V_j\gets B+1$
				\EndIf
			
				\EndIf
				\State $P\gets P-1$, $i\gets i+1$
				\EndFor
				\EndWhile
				\State $iter\gets iter + 1$
				\State $\beta\gets (iter/K)\lambda$
				\Until{$\beta > \lambda$}
				
				\For{$i=1 \to P$}
				\ForAll{$j\in G_i$}
				\State $S_j \gets V_i$
				\EndFor
				\EndFor
				\Ensure filtered Image $S$
			\end{algorithmic}
			\end{algorithm}
	\section{Experiments and Results}
	\label{sec5}
	\begin{figure*}[t]
		\centering
		\begin{minipage}[t]{0.4\linewidth}
			\centering
			\includegraphics[width=2.5in]{final/Adirondack/missing_50.png}
		\end{minipage}%
		\begin{minipage}[t]{0.4\linewidth}
			\centering
			\includegraphics[width=2.5in]{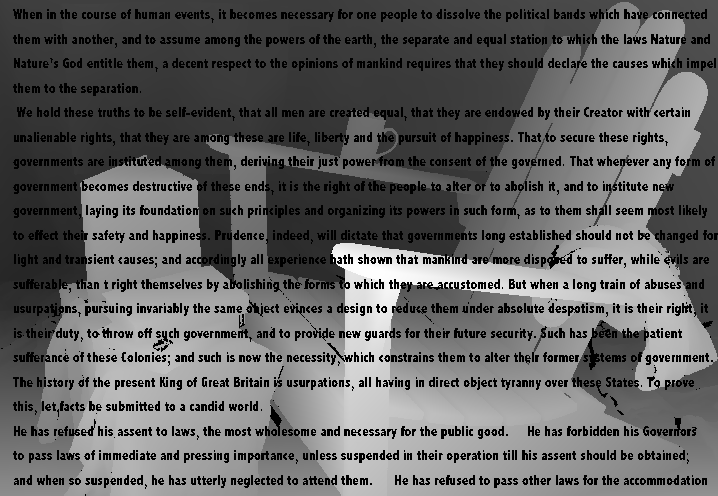}
		\end{minipage}
		\label{fig5}
	
			\caption{The left is an example of the random missing pattern. The right is an example of the textual missing mask.}
		
	\end{figure*}
	
	\subsection{Dataset} For there is no public dataset that aims at inpainting depth images, we make a dataset for evaluating depth inpainting approaches. We convert the groundtruth disparity maps from Middlebury stereo dataset \cite{scharstein2002taxonomy} to grayscale depth images. The groundtruth depth maps are from 14 images including Adirondack, Jadeplant, Motorcycle, Piano, Playtable, Playroom, Recycle, Shelves, Teddy, Pipes, Vintage, MotorcycleE, PianoL and PlaytableP. The unknown values of the groundtruth disparity maps are converted to 0s in depth images. To create damaged images, we generate several masks including random missing masks (see Figure \ref{fig5}) and textual masks (see Figure \ref{fig5}). The masks are provided in our dataset.
	
	In addition to the depth images and masks, we also provide our results which will be displayed in this section together with our codes for all the algorithms concerned in this paper. Our codes also enable generations of new random masks. New depth images can also be processed by our codes to produce inpainted results. Our dataset can be accessed via \url{http://www.cad.zju.edu.cn/home/dengcai/Data/depthinpaint/DepthInpaintData.html}. Our code can be accessed via \url{https://github.com/xuehy/depthInpainting}.
	
	\subsection{Experiments}
	\begin{figure*}[t]
		\centering
		\begin{minipage}[t]{0.33\linewidth}
			\centering
			\includegraphics[width=2in]{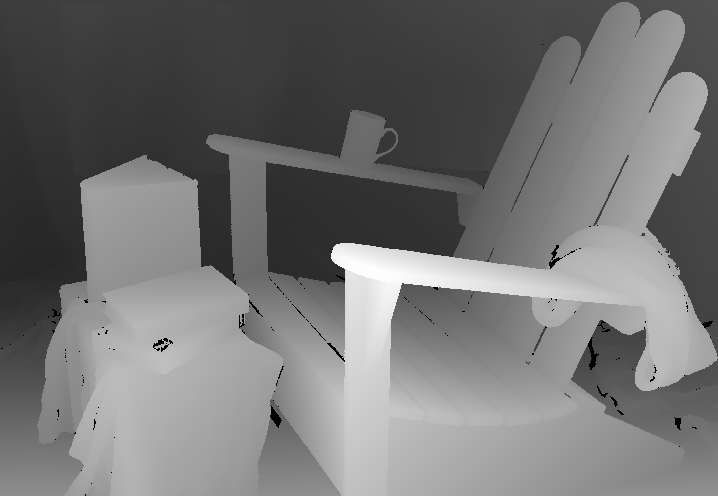}
		\end{minipage}%
		\begin{minipage}[t]{0.33\linewidth}
			\centering
			\includegraphics[width=2in]{final/adi/missing.png}
		\end{minipage}%
		\begin{minipage}[t]{0.33\linewidth}
			\centering
			\includegraphics[width=2in]{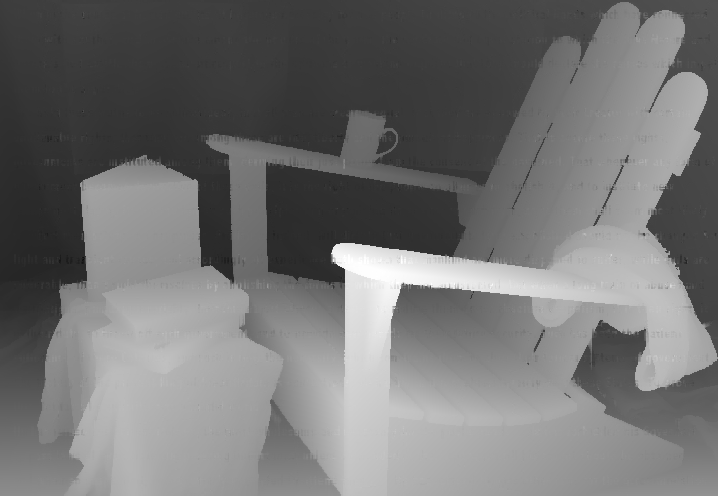}
		\end{minipage}
		\begin{minipage}[t]{0.33\linewidth}
			\centering
			\includegraphics[width=2in]{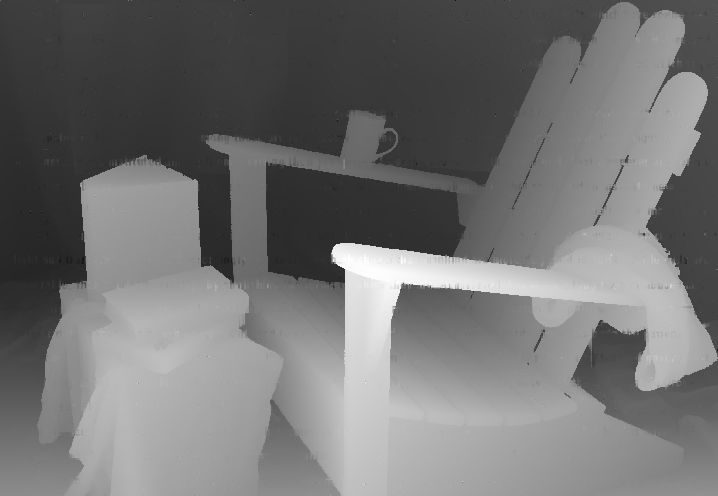}
		\end{minipage}%
		\begin{minipage}[t]{0.33\linewidth}
			\centering
			\includegraphics[width=2in]{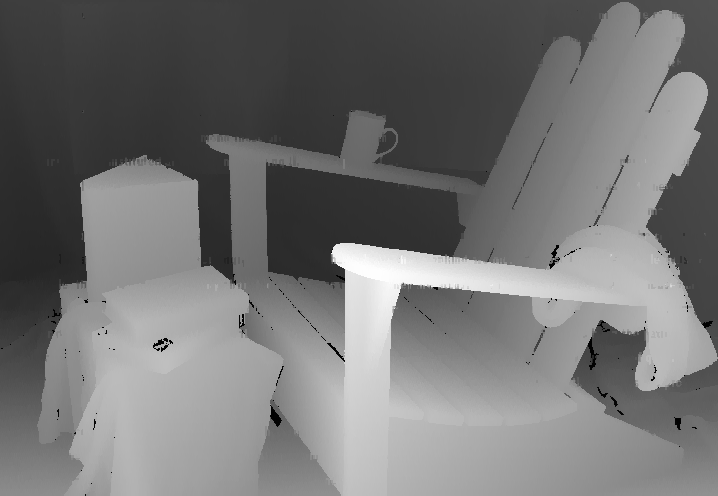}
		\end{minipage}%
		\begin{minipage}[t]{0.33\linewidth}
			\centering
			\includegraphics[width=2in]{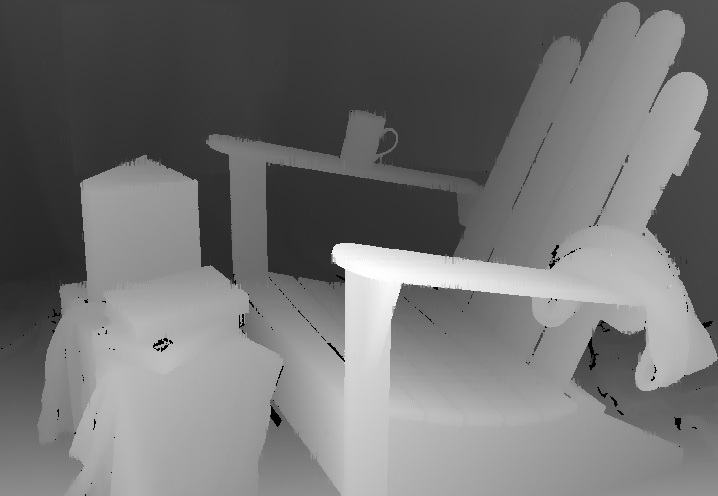}
		\end{minipage}

		\begin{minipage}[t]{0.33\linewidth}
			\centering
			\includegraphics[width=2in]{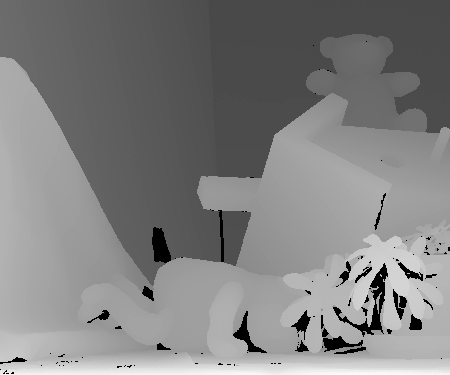}
		\end{minipage}%
		\begin{minipage}[t]{0.33\linewidth}
			\centering
			\includegraphics[width=2in]{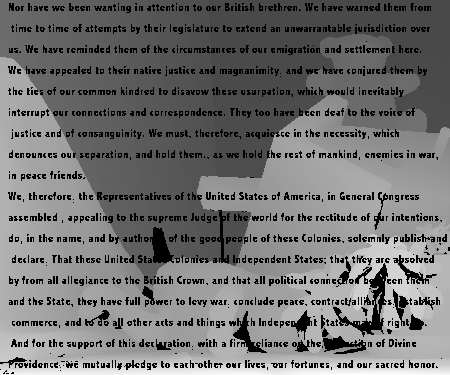}
		\end{minipage}%
		\begin{minipage}[t]{0.33\linewidth}
			\centering
			\includegraphics[width=2in]{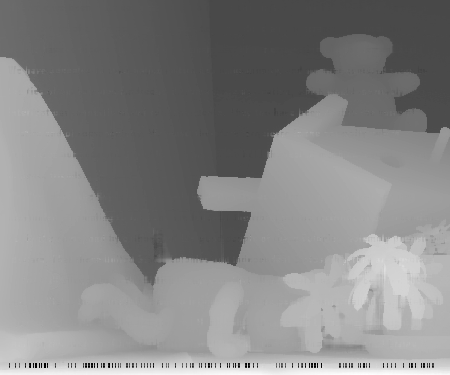}
		\end{minipage}
		\begin{minipage}[t]{0.33\linewidth}
			\centering
			\includegraphics[width=2in]{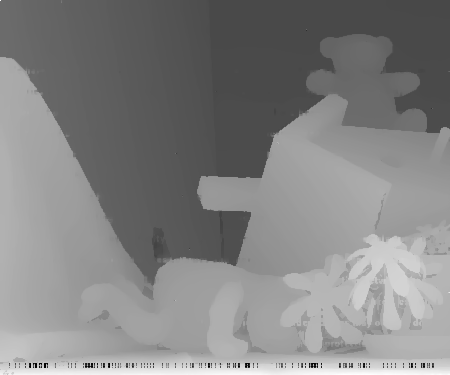}
		\end{minipage}%
		\begin{minipage}[t]{0.33\linewidth}
			\centering
			\includegraphics[width=2in]{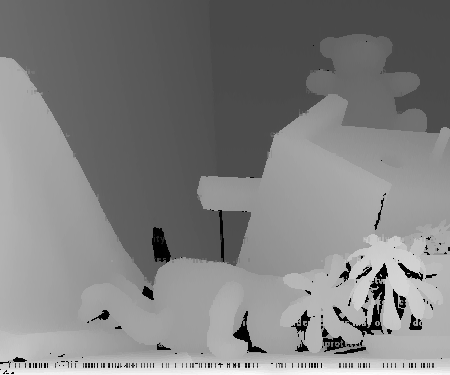}
		\end{minipage}%
		\begin{minipage}[t]{0.33\linewidth}
			\centering
			\includegraphics[width=2in]{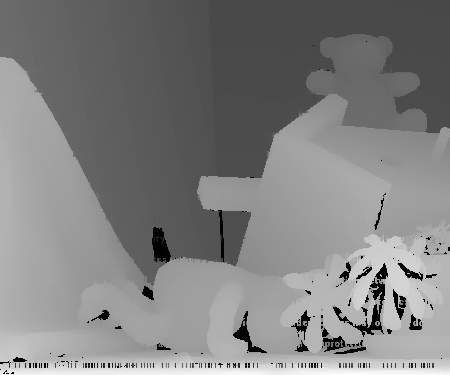}
		\end{minipage}
		
		\caption{In this figure we show the results of Adirondack and Teddy with textual mask. For every image, the first row shows the original depth image, the damaged image and the low rank result (from left to right). The second row displays the result of LRTV, LRL0 and LRL0$^\psi$ (from left to right). The PSNR of Adirondack are 28.2444(Low rank), 28.3886(LRTV), 28.9374(LRL0) and \textbf{29.2523}(LRL0$^\psi$). The PSNR of Teddy are 16.3590, 16.9535, 17.0616 and \textbf{17.1101}. }
		\label{fig6}
	\end{figure*}
		
	We first inpaint the depth images with the nuclear norm regularization matrix completion approach. Then we apply the LRTV approach which is first used by Shi \textit{et al.} \cite{shi2013low} for medical image super-resolution. From the inpainted results, we can see LRTV reduces noise caused by the low rank completion. Later on, we employ the LRL0 algorithm. As we can see, although LRL0 outperforms LRTV in most cases, it fails on some depth images. As discussed in section \ref{sec1}, we propose the low gradient measure and the corresponding LRL0$^\psi$ algorithm based on the statistics of depth gradient maps. In the end we inpaint the images with LRL0$^\psi$. We evaluate the inpainted results using PSNR and the PSNR is computed only on the missing area of the depth images.

	\subsection{Parameters and Setup}
		\begin{figure}[t]
			\centering
			\includegraphics[width=3.5in]{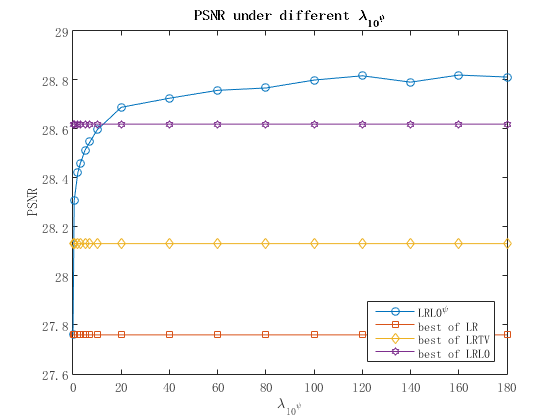}
			\caption{For the LRL0$^\psi$ algorithm we search for the best value of $\lambda_{l0^\psi}$ for the image Adirondack. The $\lambda_{l0^\psi}$ with highest PSNR is reported as the result. This figure shows the results of LRL0$^\psi$ under different parameters. The best results of the other methods are reported here. We can see our LRL0$^\psi$ achieves better results than the other approaches when the parameter gets large enough.}
			\label{fig11}
		\end{figure}
	The experiments are performed on a PC of Intel i7
	3.5GHz CPU with 16GB memory. Our parameter settings
	are listed as follows. For all algorithms, we set the weight for the low rank term $\lambda_r = 10$.
	\begin{itemize}
		\item For LRTV: we set the weight for TV term to $\lambda_{tv} = 40$.
		\item For LRL0: we set the weight as  $\lambda_{l0} = 30$.
		\item For LRL0$^\psi$: we set the weight as $\lambda_{l0^\psi} = 100$.
	\end{itemize}

	For all algorithms, the iterations are stopped when the the number of iterations exceeds 30 or the relative error of solutions is below $1\times 10^{-3}$. 
	
	\begin{figure*}[t]
		\centering
		\begin{minipage}[t]{0.33\linewidth}
			\centering
			\includegraphics[width=2in]{final/Adirondack/disp.png}
		\end{minipage}%
		\begin{minipage}[t]{0.33\linewidth}
			\centering
			\includegraphics[width=2in]{final/Adirondack/missing_50.png}
		\end{minipage}%
		\begin{minipage}[t]{0.33\linewidth}
			\centering
			\includegraphics[width=2in]{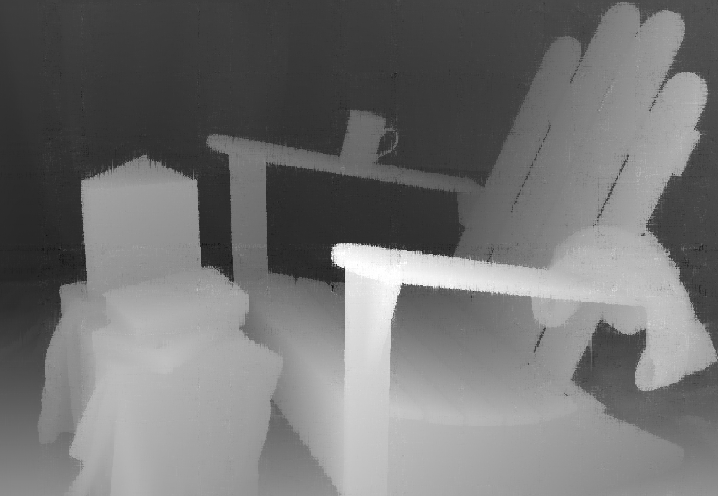}
		\end{minipage}
		\begin{minipage}[t]{0.33\linewidth}
			\centering
			\includegraphics[width=2in]{final/Adirondack/lrtv_50.png}
		\end{minipage}%
		\begin{minipage}[t]{0.33\linewidth}
			\centering
			\includegraphics[width=2in]{final/Adirondack/lrl0_50_30.png}
		\end{minipage}%
		\begin{minipage}[t]{0.33\linewidth}
			\centering
			\includegraphics[width=2in]{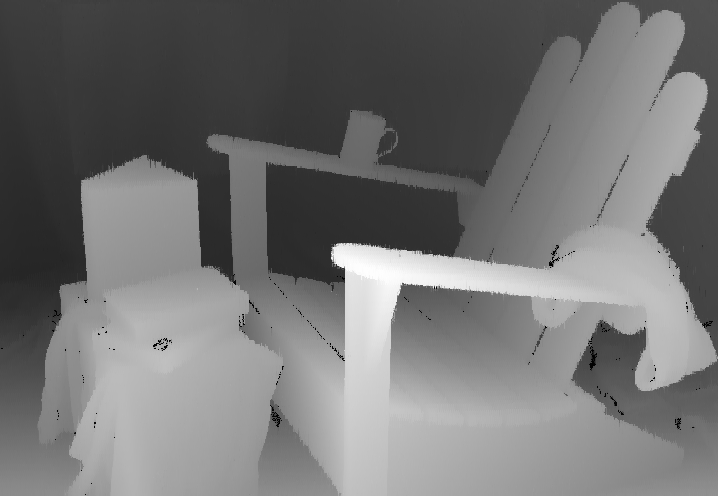}
		\end{minipage}
		\begin{minipage}[t]{0.33\linewidth}
			\centering
			\includegraphics[width=2in]{final/Jadeplant/disp.png}
		\end{minipage}%
		\begin{minipage}[t]{0.33\linewidth}
			\centering
			\includegraphics[width=2in]{final/Jadeplant/missing_50.png}
		\end{minipage}%
		\begin{minipage}[t]{0.33\linewidth}
			\centering
			\includegraphics[width=2in]{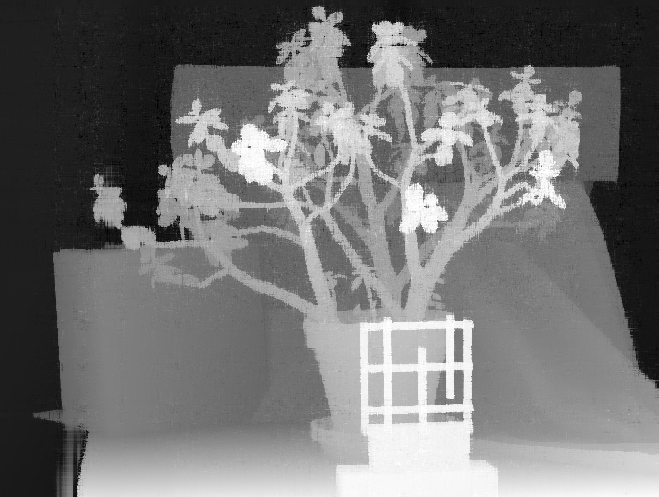}
		\end{minipage}
		\begin{minipage}[t]{0.33\linewidth}
			\centering
			\includegraphics[width=2in]{final/Jadeplant/lrtv_50.png}
		\end{minipage}%
		\begin{minipage}[t]{0.33\linewidth}
			\centering
			\includegraphics[width=2in]{final/Jadeplant/lrl0_50_30.png}
		\end{minipage}%
		\begin{minipage}[t]{0.33\linewidth}
			\centering
			\includegraphics[width=2in]{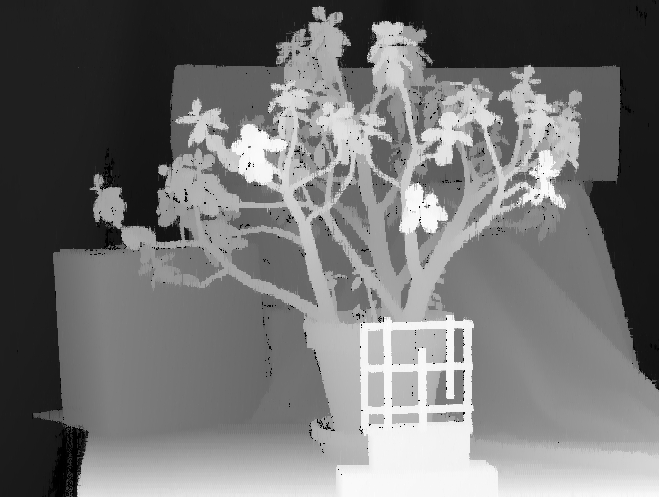}
		\end{minipage}
		\begin{minipage}[t]{0.33\linewidth}
			\centering
			\includegraphics[width=2in]{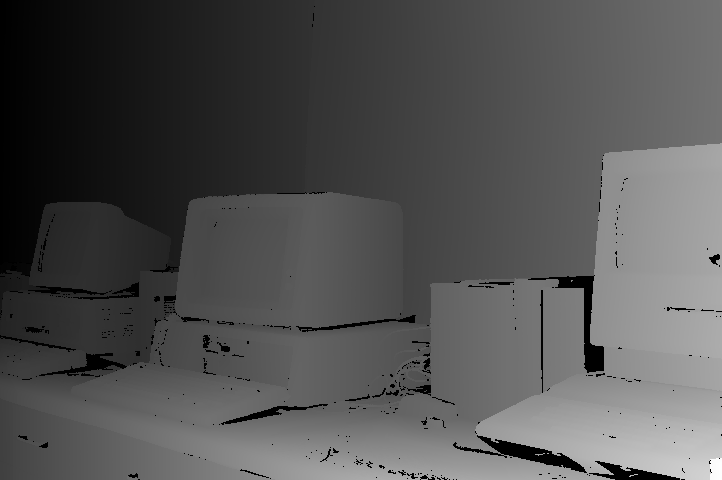}
		\end{minipage}%
		\begin{minipage}[t]{0.33\linewidth}
			\centering
			\includegraphics[width=2in]{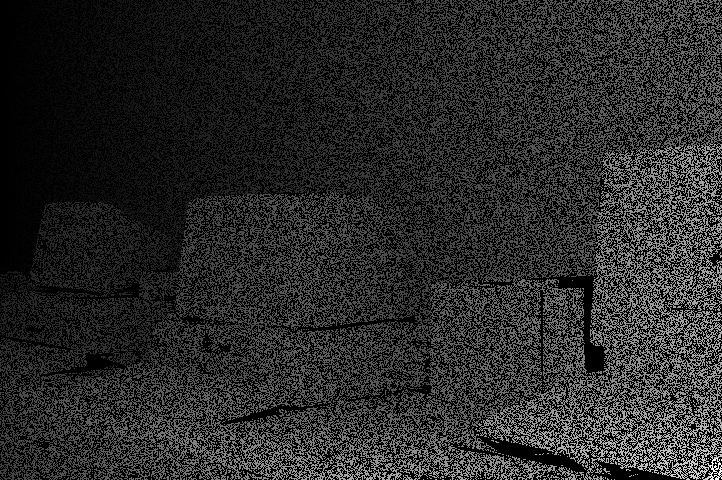}
		\end{minipage}%
		\begin{minipage}[t]{0.33\linewidth}
			\centering
			\includegraphics[width=2in]{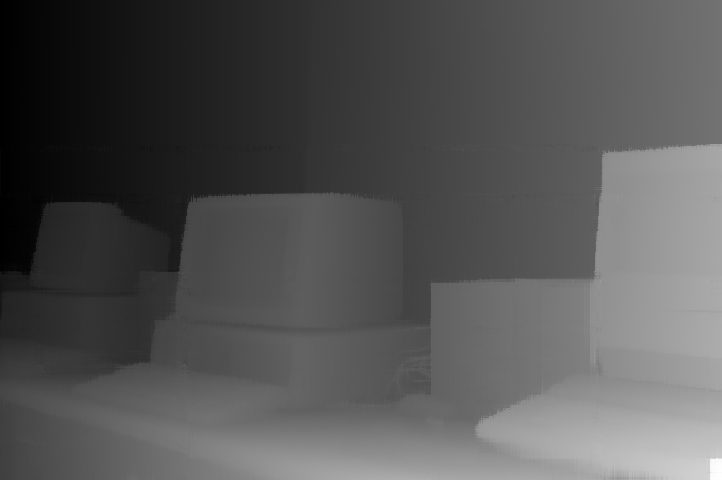}
		\end{minipage}
		\begin{minipage}[t]{0.33\linewidth}
			\centering
			\includegraphics[width=2in]{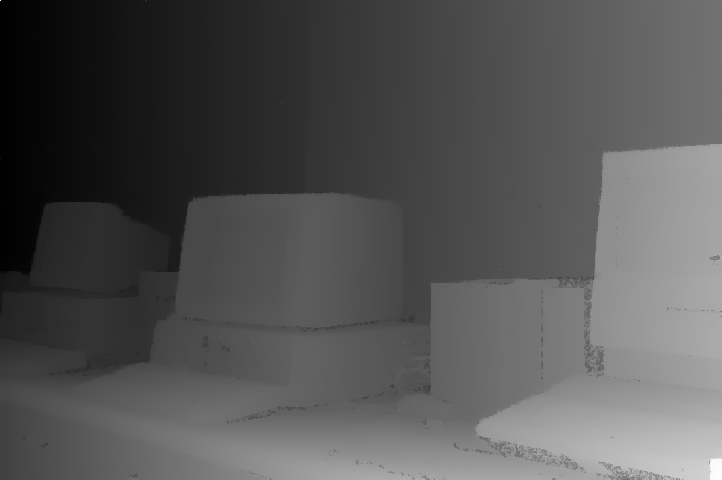}
		\end{minipage}%
		\begin{minipage}[t]{0.33\linewidth}
			\centering
			\includegraphics[width=2in]{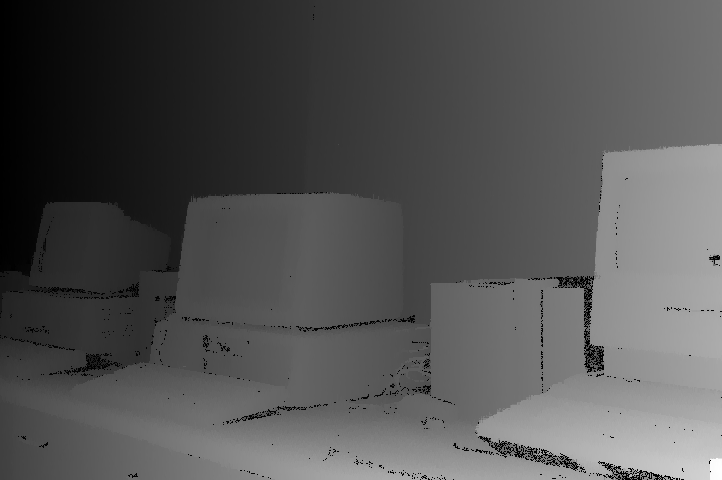}
		\end{minipage}%
		\begin{minipage}[t]{0.33\linewidth}
			\centering
			\includegraphics[width=2in]{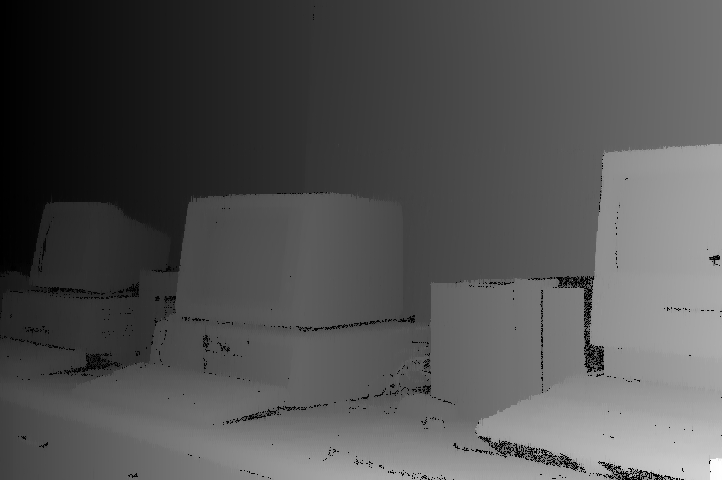}
		\end{minipage}%
		
		\caption{In this figure we show the results of three images: Adirondack, Jadeplant and Vintage. For every image, the first row shows the original depth image, the damaged image and the low rank result (from left to right). The second row displays the result of LRTV, LRL0 and LRL0$^\psi$ (from left to right). The PSNR of Adirondack are 27.7596(Low rank), 28.1319(LRTV), 28.6198(LRL0) and \textbf{28.8292}(LRL0$^\psi$). The PSNR of Jadeplant are 21.9843, 22.8231, 22.7730 and \textbf{22.8725}. The PSNR of Vintage are 22.7924, 23.2368, 24.0400 and \textbf{24.2091}. }
		\label{fig7}
	\end{figure*}
	
	\subsection{Results and Analysis}
	
	\begin{table}
		\begin{center}
			\begin{tabular}{|l|c|c|c|c|}
				\hline
				Mask type & Adiron & Jadepl & Motor & Piano \\
				\hline
				Method & rand & rand & rand & rand \\
				\hline
				LR  & 27.7596 & 21.9843 & 22.0838 & 18.2673 \\
				LRTV & 28.1319 & 22.8231 & 22.7546 & 18.5161 \\
				LRL0 & 28.6198 & 22.7730 & 22.9303 & 19.5263 \\
				LRL0$^\psi$ & \textbf{28.8292} & \textbf{22.8725} & \textbf{23.1480} & \textbf{19.6801}\\
				\hline
				\hline
				Mask type & Playt & Playrm & Recyc & Shelvs \\
				\hline
				Method & rand & rand & rand & rand \\
				\hline
				LR & 22.8384 & 17.7783 & 23.5558 & 14.9259\\
				LRTV & 23.2150 & 18.0073 & 23.9033 & 15.3300\\
				LRL0 & 24.3513 & 18.6153 & 24.7019 & 16.1661\\
				LRL0$^\psi$ & \textbf{24.4886} & \textbf{18.7681} & \textbf{24.8112} & \textbf{16.2910}\\
				\hline
				\hline
				Mask type & Teddy & Pipes & Vintge & MotorE \\
				\hline
				Method & rand & rand & rand & rand \\
				\hline
				LR & 19.5329 & 22.1899 & 22.7924 & 22.3541\\
				LRTV & 20.0866 & 22.9039 & 23.2368 & 23.2859 \\
				LRL0 & 20.7570 & 23.4278 & 24.0400 & 23.1577\\
				LRL0$^\psi$ & \textbf{20.8839} & \textbf{23.6760} & \textbf{24.2091} & \textbf{23.3850} \\
				\hline
				\hline
				Mask type & PianoL & PlaytP & adi & ted \\
				\hline
				Method & rand & rand & text & text \\
				\hline
				LR & 19.8488 & 23.9959 & 28.2444 & 16.359\\
				LRTV & 20.1330 & 23.2150 & 28.3886 & 16.9535\\
				LRL0 & 21.3011 & 25.6337 & 28.9347 & 17.0616\\
				LRL0$^\psi$ & \textbf{21.4553} & \textbf{25.8008} & \textbf{29.2523} & \textbf{17.1101} \\
				\hline
			\end{tabular}
		\end{center}
		\caption{PSNR of results from different inpainting methods on our dataset. The mask type indicates what kind of mask is applied to make the depth image with missing areas. \textit{rand} indicates random missing and \textit{text} indicates textual masks. Noting that our proposed  LRL0 approach performs generally better than LRTV and achieves relatively close results to LRL0$^\psi$. However, LRL0 gets worse-than-LRTV results on MotorE and Jadeplant. Our proposed LRL0$^\psi$ approach always achieves the best results in PSNR. }
		\label{table1}
	\end{table}
	Some of the experimental results are shown in Figure \ref{fig6} and Figure \ref{fig7}. The PSNR of the results of the whole dataset are shown in Table \ref{table1}. The low rank results have obvious noise (see Figure \ref{fig6} and \ref{fig7}). Noting that the LRL0 approach performs generally better than LRTV and achieves relatively close results to LRL0$^\psi$. However, LRL0 gets \textit{worse-than-LRTV} results on MotorE and Jadepl.  For high resolution results, please refer to our published dataset. Our proposed LRL0$^\psi$ approach always achieves the best results in PSNR because it allows for gradual pixel value variation which is common in depth images.
	
	\subsection{Parameter Analysis}
		We tune the parameter $\lambda_{l0^\psi}$ to see the effect of the weight of the low gradient regularization term. The resultant PSNR curve is shown in Figure \ref{fig11}. When $\lambda_{l0^\psi} = 0$, the LRL0$^\psi$ algorithm degenerates to the LR algorithm. As $\lambda_{l0^\psi}$ increases, the low gradient term gets more important and the inpainting result improves over the LR algorithm. When $\lambda_{l0^\psi}$ gets large enough(exceeds 80), the resultant PSNR gets stable. The LRL0$^\psi$ algorithm outperforms the other approaches when the weight $\lambda_{l0^\psi}$ for the low gradient term gets large enough. 
	\section{Conclusions} \label{sec6}
	We consider the problem of inpainting depth images. The popular image inpainting technique \textit{low rank matrix completion} always leads to noisy depth inpainting results. Based on the gradient statistics of depth images, we propose the low gradient regularization and combine it with the low rank prior into the LRL0$^\psi$ approach.   Then we extend the region fusion approach in \cite{nguyen2015fast} for our $L_0^\psi$ gradient minimization problem. 
	We compare our approach with the only low rank regularization and two other approaches, the LRTV and the LRL0 approaches, which only enforce sparse gradient regularization.
	Experiments show that our proposed methods outperform the only low rank method. Our proposed  LRL0$^\psi$ approach also outperforms the LRTV and LRL0 approaches for it better utilizes the gradient property.
	
	{\small
		\bibliographystyle{ieee}
		\bibliography{lrl0}
	}
	
	\begin{IEEEbiography}[{\includegraphics[width=1in,height=1.25in,clip,keepaspectratio]{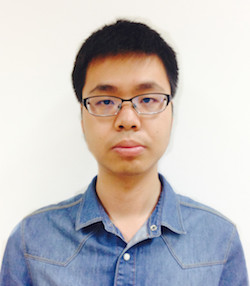}}]{Hongyang Xue}
		Hongyang Xue received the B.E. degree in Computer Science and Technology from Zhejiang University, China, in 2014. He is currently a Ph.D. candidate in Computer Science at Zhejiang University. His research include machine learning, computer vision and data mining.

	\end{IEEEbiography}
	
	\begin{IEEEbiography}[{\includegraphics[width=1in,height=1.25in,clip,keepaspectratio]{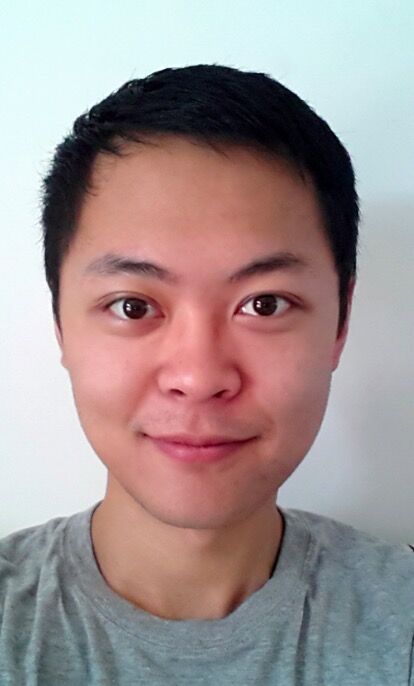}}]{Shengming Zhang}
		Shengming Zhang is an intern in the State Key Laboratory of CAD\&CG, Undergraduate of Computer Science at Zhejiang University, China. He will receive the Bachelor's degree in Computer Science from Zhejiang University in 2017. His research interests include computer vision, computer graphics and machine learning.
		
	\end{IEEEbiography}
	
	\begin{IEEEbiography}[{\includegraphics[width=1in,height=1.25in,clip,keepaspectratio]{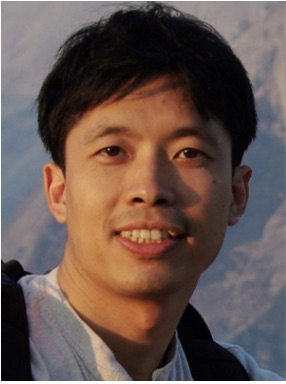}}]{Deng Cai}
		
		Deng Cai is a Professor in the State Key Laboratory of CAD\&CG, College of Computer Science at Zhejiang University, China. He received the Ph.D. degree in Computer Science from the University of Illinois at Urbana Champaign in 2009. His research interests include machine learning, data mining and information retrieval.
	\end{IEEEbiography}

\end{document}